\documentclass[letterpaper]{article} 
\usepackage{aaai2027}  
\usepackage[hyphens]{url}  
\usepackage{graphicx} 
\usepackage{natbib}  
\usepackage{caption} 
\usepackage{algorithm}
\usepackage{algorithmic}

\usepackage{newfloat}
\usepackage{amsmath}
\usepackage{amsfonts}
\usepackage{listings}
\DeclareCaptionStyle{ruled}{labelfont=normalfont,labelsep=colon,strut=off} 
\floatstyle{ruled}
\newfloat{listing}{tb}{lst}{}
\floatname{listing}{Listing}

\usepackage{booktabs}
\usepackage{multirow}

\title{MORA: Modeling Observed Changes for Drift-Robust Time-Series Anomaly Detection}
\author{
    Xudong Mou\textsuperscript{\rm 1},
    Tiejun Wang\textsuperscript{\rm 1},
    Rui Wang\textsuperscript{\rm 1},
    Hui Wang\textsuperscript{\rm 2},
    Pin Liu\textsuperscript{\rm 3},
    Tianyu Wo\textsuperscript{\rm 1},\\
    Xudong Liu\textsuperscript{\rm 1},
    Renyu Yang\textsuperscript{\rm 1}\corresponding
}
\affiliations{
     \textsuperscript{\rm 1}Beihang University, Beijing, China\\
    \textsuperscript{\rm 2}Beihang Hangzhou Innovation Institute, Hangzhou, China\\
    \textsuperscript{\rm 3}China University of Geosciences (Beijing), Beijing, China\\
    mxd@buaa.edu.cn, wtj@buaa.edu.cn, ruiking@buaa.edu.cn, whui@buaa.edu.cn,
    liupin@cugb.edu.cn, woty@buaa.edu.cn, liuxd@buaa.edu.cn, renyuyang@buaa.edu.cn
}

\begin{document}

\maketitle

\begin{abstract}
Time-series anomaly detection (TSAD) identifies deviations from patterns learned from historical data. In non-stationary settings, distribution drift and true anomalies can cause similar local changes, making it difficult to tell whether a deviation reflects abnormality or evolving context. Existing methods typically adapt to detected shifts or learn drift-insensitive representations, but do not resolve this ambiguity. We define this problem as \emph{temporal change disambiguation}: determining whether a local deviation is explained by broader temporal evolution.  
We introduce MORA, a drift-robust TSAD framework that reconstructs the same local target from paired short- and long-term views. The reconstruction gap measures contextual support for a local deviation, and a data-dependent correction mechanism conservatively adjusts the primary local anomaly score. Context can only reduce the score when it improves reconstruction of the same target. MORA needs neither drift annotations nor online adaptation. Experiments on four TSAD benchmarks show strong robustness to non-stationarity while preserving sensitivity to genuine anomalies.
\end{abstract}


\section{1. Introduction}

Time-series anomaly detection (TSAD) identifies observations that significantly deviate from normal temporal behavior~\cite{grubbs1969procedures} and is critical in industrial monitoring, AIOps, cyber-physical systems, and other safety-critical domains. Because anomalies are rare and costly to label, TSAD is usually treated as an unsupervised problem~\cite{pang2021deep}: models learn normal patterns from unlabeled or normal-only data and flag violations as anomalies. Normality is typically modeled via reconstruction~\cite{li2026crossad,malhotra2016lstm}, one-class representation learning~\cite{zhong2025simad,xu2024calibrated,ruff2018deep}, or contrastive objectives~\cite{yang2023dcdetector,sohn2020learning}. Accordingly, anomalies are typically characterized by large reconstruction/prediction errors, low likelihood, or deviations from compact normal representations. Despite their differences, these methods rely on a shared premise: normal patterns learned from the past remain valid at test time.

This premise is frequently violated in real-world time series, where normal conditions drift due to seasonality, workload fluctuations, sensor aging, maintenance, or upgrades. Such evolution changes both observation marginals and temporal dependencies, so data from a drifting normal process can differ from historical normality and receive high anomaly scores. Thus drift and anomalies first appear to a detector in the same way: as changes relative to learned normality. A detector that reacts to all such changes may raise many false alarms under gradual drift or treat an abrupt regime shift as an anomaly. The challenge is therefore not just detecting change, but interpreting what it means.

\begin{figure}[t]
    \centering
    \includegraphics[width=0.95\columnwidth]{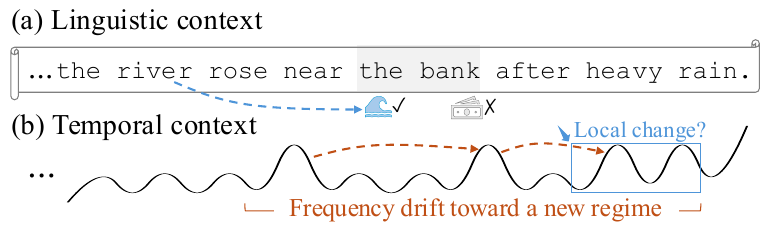}
    \caption{Motivation for temporal change disambiguation.
    (a) Linguistic context resolves the meaning of an ambiguous word.
    (b) Similarly, a local temporal change may appear anomalous in isolation,
    while its broader context reveals a coherent transition toward a new
    operating regime.}
    \label{fig:motivation}
\end{figure}

Existing studies mainly take two approaches. The first uses a detect-and-adapt paradigm~\cite{han2023anomaly}, where distribution shifts are detected and then handled via model updating, continual learning, knowledge distillation, or test-time adaptation. These methods can be effective but depend on accurate shift detection; incorrect updates may absorb anomalies into normality and add deployment cost. The second seeks drift-robust representations~\cite{wei2026evolving,huang2025graph} through temporal decomposition, multi-scale modeling, invariance, or uncertainty-aware training~\cite{lian2025facing}, 
aiming to reduce sensitivity to non-stationary changes in a single representation or score. However, it is inherently difficult to suppress drift-induced variation while preserving evidence of visually similar anomalies. As a result, many methods adapt to or suppress distributional changes without explicitly interpreting what a change represents, leaving drift–anomaly ambiguity to be resolved implicitly by the learned representation or score.

These limitations motivate us to reconsider what a detector observes under non-stationarity. As shown in Fig.~\ref{fig:motivation}, the word ``\textit{bank}'' is ambiguous in isolation but becomes clear from contextual cues like ``\textit{river}'' and ``\textit{rain}.'' Similarly, a local temporal change alone does not indicate whether it arises from an abnormal event or an evolving normal process; its meaning depends on the broader trajectory. We therefore treat a local change as ambiguous evidence rather than an anomaly by default, and refer to its contextual interpretation \emph{temporal change disambiguation}. 

Guided by this perspective, we propose MORA, a drift-robust TSAD framework that resolves locally ambiguous changes using broader temporal context. It reconstructs the same local target from paired short- and long-term views and contrasts their reconstruction errors to measure how the interpretation of a local deviation changes with additional history. It further extracts heterogeneous relations among the two views and uses them to adaptively control the strength of score correction. A conservative one-sided rule allows contextual information to reduce the primary local score only when it provides a lower reconstruction error. In this way, MORA suppresses context-supported changes while retaining deviations that remain unexplained under broader temporal context.
We summarize the main contributions as follows:
\begin{itemize}
\item We cast drift-robust TSAD as a problem of temporal change disambiguation, offering a new perspective in which local deviations are interpreted against broader temporal evolution: local observations provide the primary anomaly evidence, while temporal context supplies corrective evidence.
\item We propose context-contrasted reconstruction that reconstructs the same local target with and without extended history, using the reconstruction error reduction as contextual support to adjust the local anomaly score.
\item We design a heterogeneous cross-view router with a one-sided correction rule that suppresses only context-explainable deviations while preserving local anomaly evidence otherwise.
\end{itemize}

\section{2. Related Work}
\subsection{2.1. Time Series Anomaly Detection}

Time series anomaly detection (TSAD) seeks to identify rare departures from normal temporal behavior. Most unsupervised TSAD methods follow a \emph{normality assumption}, learning normal patterns and detecting violations via reconstruction/forecasting errors \cite{malhotra2016lstm,li2026crossad}, one-class or calibrated normality scoring \cite{ruff2018deep,zhong2025simad,xu2024calibrated}, and stronger temporal modeling for long-range and cross-variable dependencies \cite{eldele2021time,yang2023dcdetector,xie2025multivariate}. Beyond a single normal model, \emph{multi-assumption} methods combine multiple normality hypotheses (e.g., AOC~\cite{mou2023deep}, TCC~\cite{sohn2020learning}) or explicit normal--anomalous modeling (e.g., CutAddPaste~\cite{wang2024cutaddpaste}, RedLamp~\cite{obata2025robust}); multi-scale designs such as CrossAD~\cite{li2026crossad} particularly inspire us to examine drift from multiple angles. Unlike CrossAD, which reconstructs fine-grained series from downsampled coarse scales to model cross-scale associations, MORA retains the original temporal resolution and reconstructs an identical local target under nested temporal contexts. The resulting error reduction is used as corrective evidence for the primary local anomaly score.

\subsection{2.2. Drift-Robust Anomaly Detection}
Non-stationarity is a central challenge in streaming TSAD: gradual drift, seasonality, and regime shifts can inflate reconstruction/forecasting errors on normal data. Existing studies mainly tackle this issue in two ways. One line follows a detect-and-adapt paradigm, identifying distribution shifts and then updating the detector via online learning, distillation, test-time adaptation, or conservative adaptation strategies \cite{han2023anomaly,guha2016robust}. The other line aims for drift-robust scoring/representations via, e.g., invariance or uncertainty-aware training \cite{wei2026evolving,huang2025graph,lian2025facing}, calibrating scores with reference windows or robust statistics \cite{sun2023efficient}, or explicitly suppressing drift by decomposing unstable series into components \cite{wang2023drift}. However, many methods focus on adapting to or suppressing distributional changes without explicitly disambiguating drift-like changes from true anomalies. Our work addresses this ambiguity by judging how much of a local deviation is explainable under broader temporal context.

\begin{figure*}[t]
    \centering
    \includegraphics[width=0.8\linewidth]{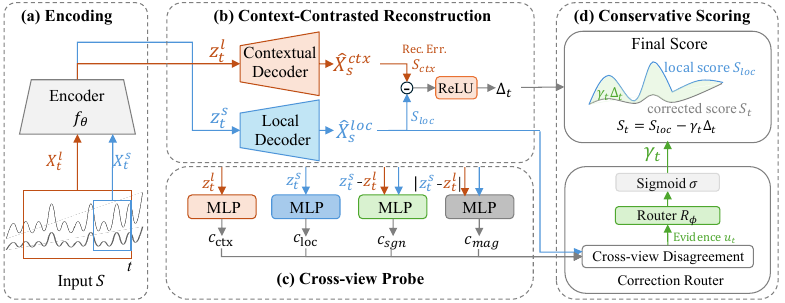}
    \caption{Overview of MORA: (a) a shared encoder for local/context windows; (b) dual reconstruction experts yield an explainable gap $\Delta_t$; (c) heterogeneous cross-view cues characterize the relation between the two views; (d) a correction router adaptively applies one-sided score correction.}
    \label{fig:framework}
\end{figure*}

\section{3. Methodology}
\subsection{3.1. Problem Definition}

Given a multivariate time series $\mathcal{S}=\{\mathbf{x}_t\}_{t=1}^{T}$ with $\mathbf{x}_t\in\mathbb{R}^{d}$, let $X_t\in\mathbb{R}^{w\times d}$ denote the local observation window at time $t$, and let $\mathcal{C}_t$ denote its broader temporal context. Time-series anomaly detection assigns an anomaly score $s_t$ to $X_t$, where a larger value indicates stronger abnormality. Most TSAD methods implicitly assume that normal patterns are stationary, such that observations can be evaluated against a time-invariant normal distribution $P_{\mathrm{ref}}(X)$. In non-stationary streams, however, the normal distribution may evolve over time. Consequently, a normal but drifting
observation can appear unlikely under $P_{\mathrm{ref}}(X)$ and receive a high anomaly score. This creates a fundamental ambiguity: both distribution drift and genuine anomalies may be manifested as deviations from previously learned normality.

The distinction lies in their temporal consistency. A drifting observation, although inconsistent with the reference distribution, may remain plausible given its broader temporal context $\mathcal{C}_t$; a genuine anomaly remains
implausible even when such context is considered. The objective is therefore to learn a drift-robust anomaly score that suppresses contextually explainable distributional changes while preserving high responses to contextually unexplained deviations.

\subsection{3.2. MORA's Approach}
\subsubsection{3.2.1 Overview.} 

The key insight behind MORA is that, under non-stationarity, local abnormality is ambiguous: a locally unusual observation may indicate either a genuine irregularity or a coherent transition of the normal process. We resolve this ambiguity through contextual explainability---a deviation that appears abnormal in a short window can be plausible when a longer temporal context is available. Accordingly, we assign asymmetric roles to the two views: the local view provides primary anomaly evidence, while the contextual view provides explanatory evidence to decide how much of that local change should be discounted.

Rather than performing independent multi-scale detection and fusing outputs, MORA evaluates the same local target under two nested information conditions
and tracks how its interpretation changes with added history. As shown in Fig.~\ref{fig:framework}, it proceeds in four stages: (i) a shared encoder maps paired short/long windows to a common latent space; (ii) two reconstruction experts score the same local target to produce local and context-conditioned abnormality evidence; (iii) a multi-view probe estimates the correction strength; and (iv) the resulting weight controls a conservative correction that removes only the explainable portion of local anomaly.

\subsubsection{3.2.2 Encoding.}
Following the standard sliding-window setting in TSAD, at each scoring step $t$ we take the local window ending at $t$ as the sample, and additionally provide a larger history window as contextual evidence:
\begin{equation}
\begin{aligned}
\mathbf{x}^{s}_t&=[\mathbf{x}_{t-L_s+1},\ldots,\mathbf{x}_{t}]\in\mathbb{R}^{L_s\times d},\\
\mathbf{x}^{l}_t&=[\mathbf{x}_{t-L_l+1},\ldots,\mathbf{x}_{t}]\in\mathbb{R}^{L_l\times d}.
\end{aligned}
\end{equation}
Here $L_l>L_s$, and $\mathbf{x}^{s}_t$ is the most recent segment of $\mathbf{x}^{l}_t$.
Both windows slide together with the same endpoint $t$; we do not run two separate detectors, but evaluate the same local target under a short-view and a context-augmented view.

A shared temporal encoder $f_{\theta}$ maps both windows into a common latent space:
\begin{equation}
\mathbf{z}^{s}_t={f}_{\theta}(\mathbf{x}^{s}_t),\qquad \mathbf{z}^{l}_t={f}_{\theta}(\mathbf{x}^{l}_t).
\end{equation}
The encoder uses temporal convolutions followed by aggregation, producing fixed-dimensional representations for variable-length inputs.
Parameter sharing ensures that differences between $\mathbf{z}^{s}_t$ and $\mathbf{z}^{l}_t$ reflect information content rather than separate encoder parameterizations.
The two representations play complementary roles: $\mathbf{z}^{s}_t$ highlights immediate behavior, making it sensitive to localized deviations, while $\mathbf{z}^{l}_t$ provides longer-term context to judge whether the same deviation is temporally explainable.

\subsubsection{3.2.3 Context-Contrasted Reconstruction.}

MORA scores the same short-term target using the two latent representations. Specifically, a local decoder and a contextual decoder reconstruct the short window as
\begin{equation}
\widehat{\mathbf{x}}^{s,\mathrm{loc}}_t=\mathcal{D}_{\mathrm{loc}}(\mathbf{z}^{s}_t),
\qquad
\widehat{\mathbf{x}}^{s,\mathrm{ctx}}_t=\mathcal{D}_{\mathrm{ctx}}(\mathbf{z}^{l}_t).
\end{equation}
We then calculate the reconstruction errors for the target $X_t^s$
\begin{equation}
\small
S^{\mathrm{loc}}_t=\frac{1}{L_s d}\left\|\mathbf{x}^{s}_t-\widehat{\mathbf{x}}^{s,\mathrm{loc}}_t\right\|_F^2,\quad
S^{\mathrm{ctx}}_t=\frac{1}{L_s d}\left\|\mathbf{x}^{s}_t-\widehat{\mathbf{x}}^{s,\mathrm{ctx}}_t\right\|_F^2.
\end{equation}
Because both decoders reconstruct the same short window $\mathbf{x}^{s}_t$, $S^{\mathrm{loc}}_t$ and $S^{\mathrm{ctx}}_t$ are directly comparable: $S^{\mathrm{loc}}_t$ measures short-view reconstruction error, while $S^{\mathrm{ctx}}_t$ measures the remaining error when longer-term context is available.
When the change at time $t$ is part of a contextually coherent transition, the local view may yield a large reconstruction error, while the contextual view can explain the deviation, resulting in $S^{\mathrm{ctx}}_t<S^{\mathrm{loc}}_t$. Conversely, if the observation is inconsistent with both the learned normality and its temporal evolution, it is difficult to reconstruct under either view and both errors remain high. Therefore, the gap between the two errors serves as a practical indicator of contextual explainability, rather than a direct estimate of drift probability.

\begin{table}[t]
\centering
\small
\renewcommand{\arraystretch}{1.0}
\begin{tabular}{lcccc}
\toprule
Metric & SMD & Exathlon & ESA & ASD \\
\midrule
Domain & Server & Spark & Satellite & Application\\
Entities & 12 & 8 & 1 & 12 \\
Train & 304168 & 88230 & 12273601 & 102331 \\
Test & 286421 & 46054 & 12162077 & 49448 \\
Dim. & 38 & 19 & 6 & 19 \\
Anomaly (\%) & 5.62 & 12.75 & 0.91 & 4.54 \\
\midrule
PSI Mean & 0.503 & 3.385 & 0.402 & 0.167 \\
JSD Mean & 0.042 & 0.212 & 0.046 & 0.016 \\
PSI Level & large & large & large & moderate \\
\bottomrule
\end{tabular}
\caption{Statistics of the four benchmark datasets.}
\label{tab:dataset_statistics}
\end{table}
\subsubsection{3.2.4 Cross-View Probe.}
A lower contextual reconstruction error does not necessarily imply a reliable explanation. The long-term window may contain irrelevant history, mixed operating regimes, or abrupt shifts that are only weakly related to the current observation. Over-trusting the contextual branch can therefore mask genuine anomaly evidence. To address this, MORA compares the long- and short-term latents from multiple complementary views. Specifically, we define
\begin{equation}
\begin{aligned}
\mathbf{v}^{\mathrm{ctx}}_t &= \mathbf{z}^{l}_t, &
\mathbf{v}^{\mathrm{loc}}_t &= \mathbf{z}^{s}_t,\\
\mathbf{v}^{\mathrm{sgn}}_t &= \mathbf{z}^{s}_t-\mathbf{z}^{l}_t, &
\mathbf{v}^{\mathrm{mag}}_t &= |\mathbf{z}^{s}_t-\mathbf{z}^{l}_t|,
\end{aligned}
\end{equation}
where the views capture contextual evidence, local evidence, the signed change between them, and its magnitude. Each view is then mapped by a dedicated MLP probe to a scalar cue $c_{\mathrm{ctx}}, c_{\mathrm{loc}}, c_{\mathrm{sgn}}, c_{\mathrm{mag}}$. 
All heads are trained to predict the local reconstruction difficulty \(S_t^{loc}\). Because they operate on representations with different semantic content, their residual discrepancies describe how differently local reconstruction difficulty is expressed across temporal views. 

Let $\mathbf{c}_t=[c_{\mathrm{ctx}},c_{\mathrm{loc}},c_{\mathrm{sgn}},c_{\mathrm{mag}}]$. We summarize their dispersion through
\begin{equation}
\small
\boldsymbol{\psi}_t=
\left[
\operatorname{Std}(\mathbf{c}_t),\,
|c_{\mathrm{loc}}-c_{\mathrm{ctx}}|,\,
|c_{\mathrm{sgn}}-c_{\mathrm{loc}}|,\,
|c_{\mathrm{mag}}-c_{\mathrm{loc}}|
\right].
\end{equation}
The first term captures overall cross-view dispersion, while the remaining terms describe differences associated with context, signed change, and change magnitude. We concatenate these statistics with the primary local score, $\mathbf{u}_t=[\boldsymbol{\psi}_t,S^{\mathrm{loc}}_t]$
and feed $\mathbf{u}_t$ to a learnable correction router
\begin{equation}
\gamma_t=\sigma\!\left(\mathcal{R}_{\phi}(\mathbf{u}_t)\right),
\qquad \gamma_t\in[0,1],
\end{equation}
where \(\gamma_t\) is a data-dependent correction weight that controls how much of the positive reconstruction gap is subtracted from the local score.

\subsubsection{3.2.5 Conservative Scoring.}

We quantify the reconstruction advantage provided by the contextual view as
\begin{equation}
\Delta_t=\left[S^{\mathrm{loc}}_t-S^{\mathrm{ctx}}_t\right]_+,
\qquad [a]_+=\max(a,0).
\end{equation}
We interpret $\Delta_t$ as the potentially explainable part of the local discrepancy. The positive-part operator enforces an asymmetric rule: contextual evidence is regarded as explanatory only when it reduces the reconstruction error for the same target.
The final anomaly score is then defined as
\begin{equation}
S_t=S^{\mathrm{loc}}_t-\gamma_t\Delta_t,
\end{equation}
which grants the contextual branch permission only to ``remove'' anomaly evidence, but never to add it. Concretely, if $S^{\mathrm{ctx}}_t\geq S^{\mathrm{loc}}_t$, then $\Delta_t=0$ and the local score is kept unchanged. If $S^{\mathrm{ctx}}_t<S^{\mathrm{loc}}_t$, only an adaptively weighted fraction of the explainable discrepancy is subtracted.
The resulting score is bounded by
$\min\left(S^{\mathrm{loc}}_t,S^{\mathrm{ctx}}_t\right)
\leq S_t\leq S^{\mathrm{loc}}_t$.

\subsubsection{3.2.6 Learning Objective.}
MORA is optimized without point-wise anomaly or drift supervision. The objective combines three components: optimizing the final drift-robust score on reference data, preserving the reconstruction behavior of both experts, and supervising the heterogeneous probes so that their disagreement becomes informative for the correction routing.
The complete objective is
\begin{equation}
\small
\begin{aligned}
\mathcal{L}
={}&
\frac{1}{N}\sum_{t=1}^{N}S_t
+
\frac{1}{N}\sum_{t=1}^{N}
\left(
 S^{\mathrm{loc}}_t+
\lambda S^{\mathrm{ctx}}_t
\right)
+\eta\mathcal{L}_{\mathrm{probe}},
\end{aligned}
\end{equation}
where the first term learns low final scores on reference data, the second regularizes the reconstruction discrepancies of the local and contextual experts, and the third trains the probe heads. Hyperparameters $\lambda$ and $\eta$ control the contributions of the contextual reconstruction and probe losses, respectively.

The probe loss is defined as
\begin{equation}
\mathcal{L}_{\mathrm{probe}}
=
\frac{1}{4N}
\sum_{t=1}^{N}\sum_{k=1}^{4}
\left(
c^{(k)}_t-
\operatorname{sg}(S^{\mathrm{loc}}_t)
\right)^2,
\end{equation}
where \(\operatorname{sg}(\cdot)\) denotes stop-gradient. The four heads receive different latent views but predict the same local reconstruction difficulty. This shared target aligns their output scales, while the remaining cross-view differences provide heterogeneous descriptors for the correction router. The router is optimized jointly with the reconstruction model through the final corrected score.

We additionally consider an optional balance regularizer,
\begin{equation}
\mathcal{L}_{\mathrm{bal}}
=
\rho
\left(
\frac{1}{|\mathcal{B}|}
\sum_{t\in\mathcal{B}}\gamma_t-\tau
\right)^2,
\end{equation}
where \(\tau\) is the target batch-average correction weight and \(\rho\) controls the regularization strength, $\mathcal{B}$ represents the mini batch. It stabilizes the average routing level during training. Since \(\rho\) is zero or small in the final configurations, we omit \(\mathcal L_{\mathrm{bal}}\) from the main objective for clarity.

During inference, the continuous score $S_t$ is used for anomaly ranking or threshold-based detection. Pseudocode is provided in Appendix~A.

\begin{table*}[t]
\centering
\small
\setlength{\tabcolsep}{3.6pt}
\renewcommand{\arraystretch}{1.0}
\begin{tabular}{lccccccccccccc}
\toprule
& \multicolumn{3}{c}{\textbf{SMD}} & \multicolumn{3}{c}{\textbf{Exathlon}} & \multicolumn{3}{c}{\textbf{ESA}} & \multicolumn{3}{c}{\textbf{ASD}} &   \\
\cmidrule(lr){2-4} \cmidrule(lr){5-7} \cmidrule(lr){8-10} \cmidrule(lr){11-13}
\textbf{Method} & \textbf{RPA-F1} & \textbf{V-R} & \textbf{V-P} & \textbf{RPA-F1} & \textbf{V-R} & \textbf{V-P} & \textbf{RPA-F1} & \textbf{V-R} & \textbf{V-P} & \textbf{RPA-F1} & \textbf{V-R} & \textbf{V-P} & \textbf{Avg. Rank $\downarrow$} \\
\midrule
IF~\shortcite{liu2012isolation}        & 6.02  & 88.11 & 42.00 & 44.54 & 92.19 & 72.41 & 0.09 & 99.40& 64.73& 16.92 & 84.02 & 33.87 & 17.125 \\
DAMP~\shortcite{lu2022matrix}      & 4.29  & 50.72 & 8.08  & 16.51 & 53.26 & 23.36 & 0.09 & 50.34& 0.92 & 1.82  & 52.04 & 6.58  & 22.375 \\
RAS~\shortcite{kim2022towards}       & 8.66  & 73.92 & 18.12 & 22.80 & 81.45 & 52.20 & 1.74 & 87.66& 7.85 & 14.42 & 83.17 & 32.48 & 17.500 \\
LSTM-ED~\shortcite{malhotra2016lstm}   & 30.87 & 85.60 & 47.96 & 51.64 & 85.28 & 65.33 & 47.59 & 94.92   &  65.67  & 24.55 & 69.35 & 20.07 & 9.250 \\
Deep SVDD~\shortcite{ruff2018deep} & 48.20 & 94.72 & 62.59 & 55.79 & 94.14 & 76.62 & 4.05 & 96.26& 41.39& 33.01 & 87.82 & 47.25 & 8.750 \\
\midrule

AOT~\shortcite{xu2021anomaly}       & 17.41 & 92.71 & 68.39 & 14.39 & \underline{97.82} & \underline{97.01} & 1.13 & 49.26& 1.02 & 6.65  & \underline{96.64} & \textbf{93.57 }& 18.500 \\
MixMamba~\shortcite{alkilane2024mixmamba}  & 27.84 & 68.06 & 17.62 & 44.51 & 76.95 & 49.89 & 25.56& 91.22& 10.66& 31.84 & 73.71 & 32.12 & 9.750 \\
MTSCAD~\shortcite{si2023beyond}    & 51.89 & 74.31 & 19.30 & 63.31 & 79.70 & 52.09 & 51.64& 91.45& 10.78& 27.17 & 73.44 & 28.46 & 5.750 \\
SHUE~\shortcite{feng2024sensitivehue}      & 20.76 & 88.05 & 47.43 & 52.03 & 92.46 & 74.77 & 15.48& \underline{99.47}& 74.35& 12.20 & 85.73 & 37.86 & 12.250 \\
MOC~\shortcite{wang2026moc}       & 38.56 & 86.45 & 48.35 & 39.09 & 90.37 & 75.32 & 2.99 & 79.72& 15.20& 38.09 & 89.71 & 56.64 & 10.500 \\
CATCH~\shortcite{wu2024catch}     & 5.31  & 88.96 & 45.10 & 11.85 & 96.05 & 88.69 & 0.36 & 82.34& 5.19 & 10.44 & 96.31 & 83.33 & 21.250 \\
MtsCID~\shortcite{xie2025multivariate}    & 9.86  & 77.74 & 65.39 & 28.46 & 97.06 & 96.76 & 2.44 & 54.75& 10.32& 21.60 & 94.30 & 92.21 & 15.500 \\
COUTA~\shortcite{xu2024calibrated}     & 4.33  & 78.35 & 39.83 & 28.51 & 93.76 & 84.04 & 0.64 & 82.75& 4.87 & 13.98 & 89.15 & 66.46 & 18.500 \\
OMNI~\shortcite{su2019robust}      & 3.50  & 52.48 & 23.30 & 12.38 & 79.23 & 66.75 & 0.18 & 47.20& 1.60 & 12.16 & 75.50 & 46.62 & 21.750 \\
GDN3~\shortcite{deng2021graph}      & 15.44 & 73.97 & 35.16 & 21.55 & 82.95 & 51.54 & 3.67 & 25.18& 5.45 & 1.52  & 58.05 & 8.55  & 18.000 \\
MindTS~\shortcite{hu2026mindts} & 51.32 & 81.41 & 35.19 & 50.22 & 77.63 & 42.75 & 22.44 & 56.58 & 8.63 & 24.65 & 82.48 & 22.91 & 9.000 \\
TranAD~\shortcite{tuli2022tranad}    & 55.31 & 90.86 & 51.71 & 66.03 & 94.29 & 76.52 & 60.00& \textbf{99.91}& \textbf{92.94}& 36.82 & 86.31 & 34.43 & \underline{3.250} \\
D3R~\shortcite{wang2023drift}       & 49.95 & 80.06 & 46.47 & 61.37 & 50.80 & 27.22 & \textbf{88.88}& 74.21& 25.68& 3.93  & 65.86 & 13.86 & 9.000 \\
DCDetector~\shortcite{yang2023dcdetector} & 9.75  & 77.36 & 66.34 & 30.12 & 97.31 & \textbf{97.12} & 0.06 & 50.31& 2.16 & 22.10 & 94.34 & \underline{92.51} & 17.000 \\
CrossAD~\shortcite{li2026crossad} &54.13& 92.58 &59.15 &42.00 & 94.15  & 74.12 &42.55	& 97.53	& 64.87	&27.91 & 86.27 &31.03 & 8.250 \\
AOC~\shortcite{mou2023deep}       & \underline{60.74} & \underline{96.48} & \underline{75.03} & 61.64 & 96.66 & 83.80 & 55.07& 97.23& 66.42& 35.49 & 91.85 & 55.67 & 4.000 \\
TCC~\shortcite{eldele2021time}       & 9.25  & 77.35 & 29.33 & 61.05 & 89.35 & 76.65 & 12.28& 81.23& 26.82& 13.20 & 85.27 & 36.80 & 13.000 \\
CutAddPaste~\shortcite{wang2024cutaddpaste} & 8.94 & 83.97 & 44.44 & \underline{70.24} & 95.80 & 86.67 & 5.91 & 90.05& 58.99& \textbf{50.52} & 94.22 & 67.85 & 8.250 \\
\midrule
MORA      &\textbf{69.97 }& \textbf{98.81} & \textbf{89.65 } & \textbf{74.84} & \textbf{98.93 }& 94.25 & \underline{84.62}& 99.38& \underline{78.79}& \underline{42.52} & \textbf{98.27} & 85.91 & \textbf{1.500 }\\
\bottomrule
\end{tabular}
\caption{Mean Best RPA-F1 (RPA-F1), VUS-ROC (V-R), and VUS-PR (V-P) results (\%) over five runs with different random seeds. The average rank based on
Best RPA-F1 is reported in the last column, where a lower value is better.}
\label{tab:main_results}
\end{table*}

\section{4. Experiments}
\subsection{4.1. Experimental Setup}
\noindent\textbf{Datasets.}We evaluate anomaly detection on multivariate benchmarks, including SMD~\cite{su2019robust, li2021multivariate}, Exathlon~\cite{jacob2020exathlon}, ESA~\cite{kotowski2024european}, and ASD~\cite{li2021multivariate}. Dataset statistics are summarized in Table~\ref{tab:dataset_statistics}~\cite{su2019robust, jacob2020exathlon, kotowski2024european, li2021multivariate}. We use Population Stability Index (PSI) and Jensen--Shannon Divergence (JSD) to summarize the train–test distribution discrepancy, and treat PSI $> 0.25$ as large following the commonly used rule of thumb.

\noindent\textbf{Baselines.}
We compare MORA with representative baselines in Table~\ref{tab:main_results}, spanning traditional detectors, deep normality methods, and multi-assumption approaches.

\noindent\textit{(1) Traditional baselines.}
We include Isolation Forest (IF)~\cite{liu2012isolation} and DAMP~\cite{lu2022matrix}.
Following~\cite{kim2022towards}, we also report Randomized Anomaly Score (RAS), which samples a score uniformly from $[0,1]$.
For ESA, these baselines are adapted to handle long sequences.
\noindent\textit{(2) Deep normality baselines.}
We evaluate representative deep detectors under common modeling assumptions, including reconstruction/prediction models (LSTM-ED~\cite{malhotra2016lstm}, TranAD~\cite{tuli2022tranad}, Anomaly Transformer (AOT)~\cite{xu2021anomaly}, MixMamba~\cite{alkilane2024mixmamba}, MTSCAD~\cite{si2023beyond}, SHUE~\cite{feng2024sensitivehue}, OMNI~\cite{su2019robust}, COUTA~\cite{xu2024calibrated}, MtsCID~\cite{xie2025multivariate}, CATCH~\cite{wu2024catch}, MOC~\cite{wang2026moc}, CrossAD~\cite{li2026crossad}, and GDN~\cite{deng2021graph}), MindTS~\cite{hu2026mindts} as well as one-class classification via Deep SVDD~\cite{ruff2018deep}.
We additionally include drift-aware scoring with D$^3$R~\cite{wang2023drift} and DCDetector~\cite{yang2023dcdetector}.
\noindent\textit{(3) Multi-assumption baselines.}
We include AOC~\cite{mou2023deep} and TCC~\cite{sohn2020learning,eldele2021time} as fused/multi-assumption methods, and CutAddPaste~\cite{wang2024cutaddpaste} as an anomaly-assumption based baseline.


\noindent\textbf{Metrics.}
We report Best RPA-F1~\cite{hundman2018detecting} as an upper-bound estimate for all methods or baselines, and use VUS-ROC/VUS-PR~\cite{paparrizos2022volume} for threshold-agnostic evaluation. 

\noindent\textbf{Implementation.}
The models are implemented using PyTorch 2.6 and Merlion 2.0.0~\cite{bhatnagar2021merlion}, and evaluated with five independent random seeds on an NVIDIA Tesla V100 GPU. We report mean results in the main paper and provide full mean$\pm$std details in the Appendix.
\begin{table*}[t]
\centering
\label{tab:ablation}

{\setlength{\tabcolsep}{3.6pt}
\renewcommand{\arraystretch}{1.0}
\small
\begin{tabular}{@{}l*{15}{c}@{}}
\toprule
& \multicolumn{3}{c}{SMD} & \multicolumn{3}{c}{Exathlon} & \multicolumn{3}{c}{ESA} & \multicolumn{3}{c}{ASD} & \multicolumn{3}{c}{Avg.} \\
\cmidrule(lr){2-4} \cmidrule(lr){5-7} \cmidrule(lr){8-10} \cmidrule(lr){11-13} \cmidrule(lr){14-16}
& F1 & V-R & V-P & F1 & V-R & V-P & F1 & V-R & V-P & F1 & V-R & V-P & F1 & V-R & V-P \\
\midrule
w/o context & 67.16 & 98.55 & 87.14 & 74.81 & 97.88 & 89.35 & 80.68 & 99.07 & 77.92 & 42.34 & 98.27 & 85.83 & 66.25 & 98.44 & 85.06 \\
w/o local   & 51.14 & 98.37 & 89.50 & 74.77 & 97.98 & 89.54 & \textbf{88.89} & \textbf{99.47} & 77.53 & 22.24 & 97.59 & 84.77 & 59.26 & 98.35 & 85.33 \\
\midrule
Naive fusion  & 68.45 & 98.54 & 87.78 & 74.77 & 97.95 & 89.51 & 84.62 & 99.40 & \textbf{78.89} & 37.44 & 97.83 & 84.64 & 66.32 & 98.43 & 85.21 \\
\midrule
Single head & 66.19 & 98.53 & 88.17 & 74.77 & 97.91 & 89.47 & 84.62 & 99.39 & 78.64 & 40.38 & 98.03 & 85.25 & 66.49 & 98.46 & 85.39 \\
Homo head   & 64.80 & 98.49 & 88.28 & 74.74 & 97.90 & 89.49 & 84.62 & 99.38 & 78.66 & 40.28 & 98.19 & 85.66 & 66.11 & 98.49 & 85.52 \\
Full MORA     & \textbf{69.97 }& \textbf{98.81} & \textbf{89.65 }& \textbf{74.84} & \textbf{98.93} & \textbf{94.25} & 84.62 & 99.38 & 78.79 & \textbf{42.52} & \textbf{98.27} & \textbf{85.91} & \textbf{67.99 }& \textbf{98.85} & \textbf{87.15 }\\
\bottomrule
\end{tabular}
}
\caption{Ablation results of mean Best RPA-F1(\%), VUS-ROC(V-R)(\%), and VUS-PR(V-P)(\%) of 5 runs with different seeds.}
\label{tab:ablation_results}
\end{table*}

\begin{figure*}[t]
    \centering
    \begin{minipage}[t]{0.24\linewidth}
        \centering
        \includegraphics[width=\linewidth]{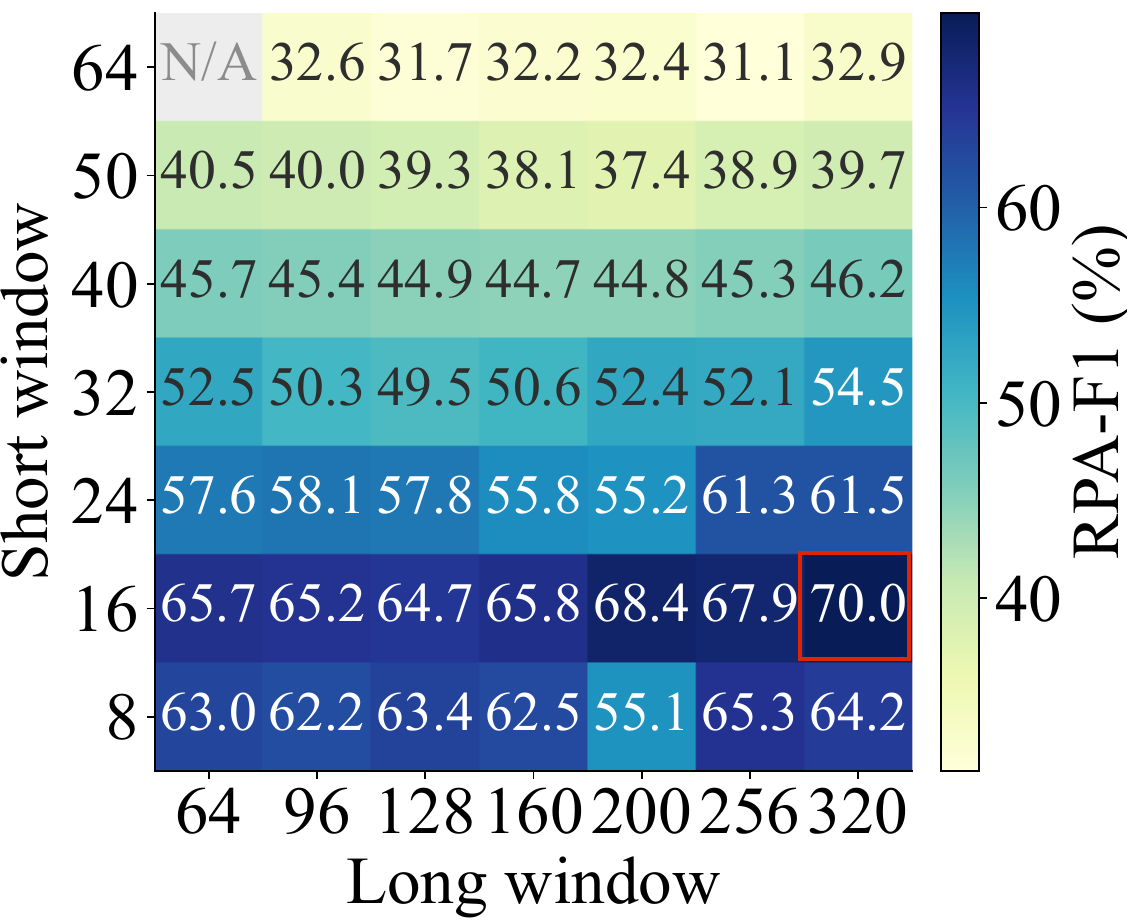}
        \caption*{(a) SMD}
    \end{minipage}\hfill
    \begin{minipage}[t]{0.24\linewidth}
        \centering
        \includegraphics[width=\linewidth]{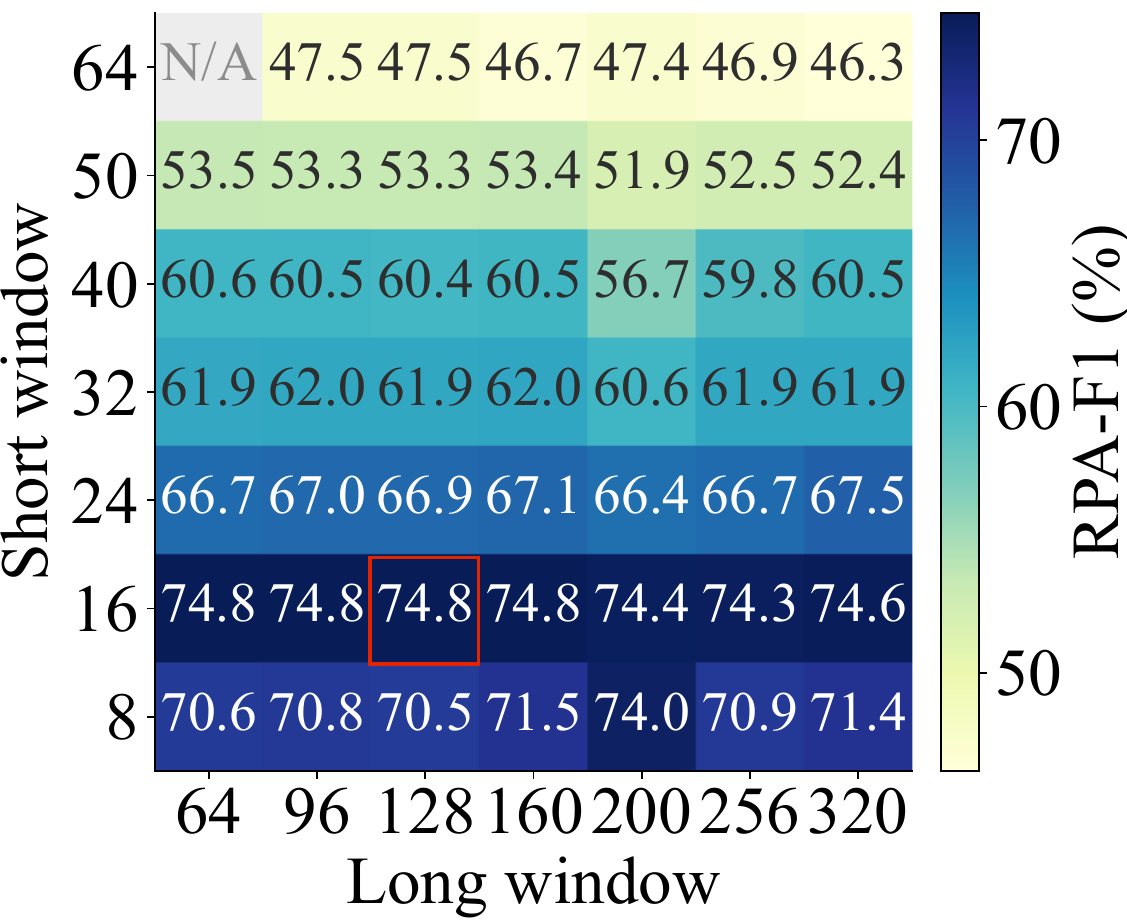}
        \caption*{(b) Exathlon}
    \end{minipage}
    \begin{minipage}[t]{0.24\linewidth}
        \centering
        \includegraphics[width=\linewidth]{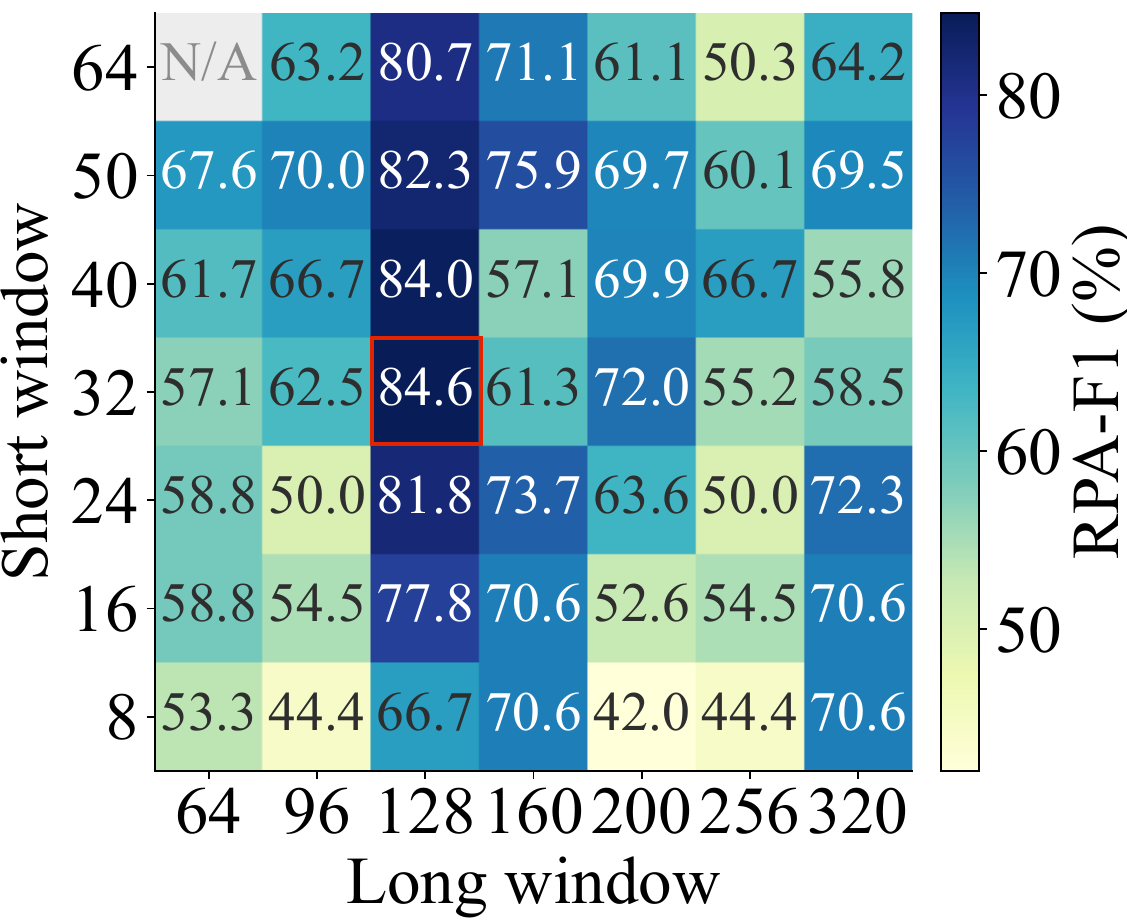}
        \caption*{(c) ESA}
    \end{minipage}\hfill
    \begin{minipage}[t]{0.24\linewidth}
        \centering
        \includegraphics[width=\linewidth]{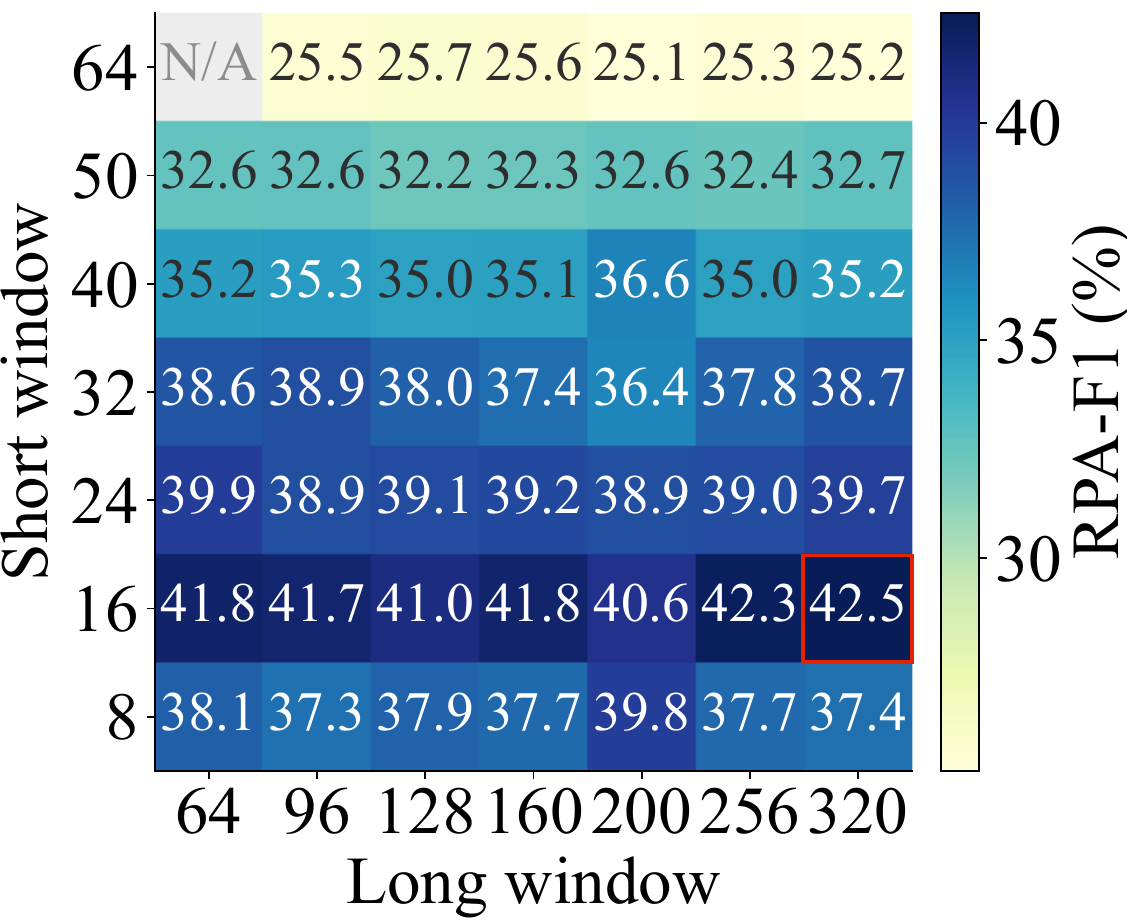}
        \caption*{(d) ASD}
    \end{minipage}

\caption{Window-pair sensitivity of RPA-F1 under different short- and long-window sizes.}
\label{fig:window_sensitivity}
\end{figure*}
\subsection{4.2. Main Results}
Table~\ref{tab:main_results} reports Best RPA-F1, VUS-ROC, and VUS-PR on four benchmarks. The last column reports the average Best RPA-F1 rank across the four datasets, following the comparison protocol of~\cite{IsmailFawaz2018deep}. Overall, MORA achieves the best average rank, placing first or second on RPA-F1 across all datasets and ranking in the top two on 9/12 dataset--metric combinations.
MORA performs best overall on SMD, leading all three metrics with clear margins, suggesting that the proposed cross-view correction effectively suppresses explainable fluctuations while preserving true abnormal events in noisy industrial telemetry. 
On Exathlon, it ranks first on RPA-F1 and VUS-ROC while remaining competitive on VUS-PR, implying a favorable trade-off between event localization and threshold robustness under diverse and frequently changing workloads. On ESA and ASD, where several baselines exhibit strong dataset-specific peaks, MORA stays near the top on both RPA-F1 and VUS metrics, indicating balanced gains in event-level accuracy and ranking quality even for long-horizon and heterogeneous temporal patterns.
These results support the effectiveness of contextual explainability and adaptive one-sided correction under diverse temporal dynamics.

\subsection{4.3. Ablation Study}
Table~\ref{tab:ablation_results} reports an ablation study of MORA. The complete model yields the best averaged performance and stays consistently strong across datasets, indicating that the proposed components are complementary rather than redundant.

\noindent\textbf{Local vs. contextual views.} Removing the local branch causes the largest degradation: the average F1 drops from $67.99\%$ to $59.26\%$, with pronounced declines on SMD and ASD. This suggests that fine-grained local evidence is crucial for capturing short-lived deviations. Notably, the context-only variant performs best on ESA, implying that long-range dependencies are more important for this ultra-long benchmark; however, its weaker results on the other datasets show that context alone is insufficient. Conversely, removing the contextual branch leads to a smaller but consistent drop, confirming that contextual cues help disambiguate locally suspicious patterns.

\noindent\textbf{Conservative correction.} Replacing the proposed one-sided correction with naive weighted fusion reduces the averaged RPA-F1 from $67.99\%$ to $66.32\%$ and VUS-PR from $87.15\%$ to $85.21\%$. The gap is most evident on Exathlon. These results highlight the benefit of assigning asymmetric roles to the two views: contextual evidence should only suppress the explainable portion of local abnormality, rather than being fused indiscriminately with the primary anomaly evidence.

\noindent\textbf{Heterogeneous probing.}
Replacing the heterogeneous probe with a single head reduces the average RPA-F1 from $67.99\%$ to $66.49\%$ and VUS-PR from $87.15\%$ to $85.39\%$. Using homogeneous heads produces a similar degradation, yielding $66.11\%$ RPA-F1 and $85.52\%$ VUS-PR. In contrast, the complete heterogeneous design achieves the best averaged result on every metric. This verifies that examining complementary local, contextual, and cross-view relations provides more informative cross-view correction cues than either a single probe or multiple
homogeneous probes.

\subsection{4.4. Sensitivity Analysis}

\noindent\textbf{Window size.} We analyze MORA's sensitivity to temporal window sizes (Fig.~\ref{fig:window_sensitivity}). Performance varies notably across short--long pairs, confirming that temporal scale is critical for disambiguating temporal changes. On SMD, the best setting
is $(L_s,L_l)=(16,320)$, while overly large $L_s$ consistently hurts, suggesting that local anomaly evidence should be captured at fine granularity and that a longer context is useful for explaining gradual changes. Exathlon shows a similar pattern, with $L_s\approx 16$ performing strongly across a wide range of $L_l$.
ESA depends more strongly on the window pair, and performs best with a
moderately large short window and sufficient context, e.g., $(32,128)$,
indicating that the appropriate scale depends on event duration. ASD is less
variable but still benefits from a proper local--context combination. Overall, the local--context separation is essential for robust anomaly scoring, rather than a minor hyperparameter choice.

\begin{figure}[t]
    \centering
    \includegraphics[width=1\columnwidth]{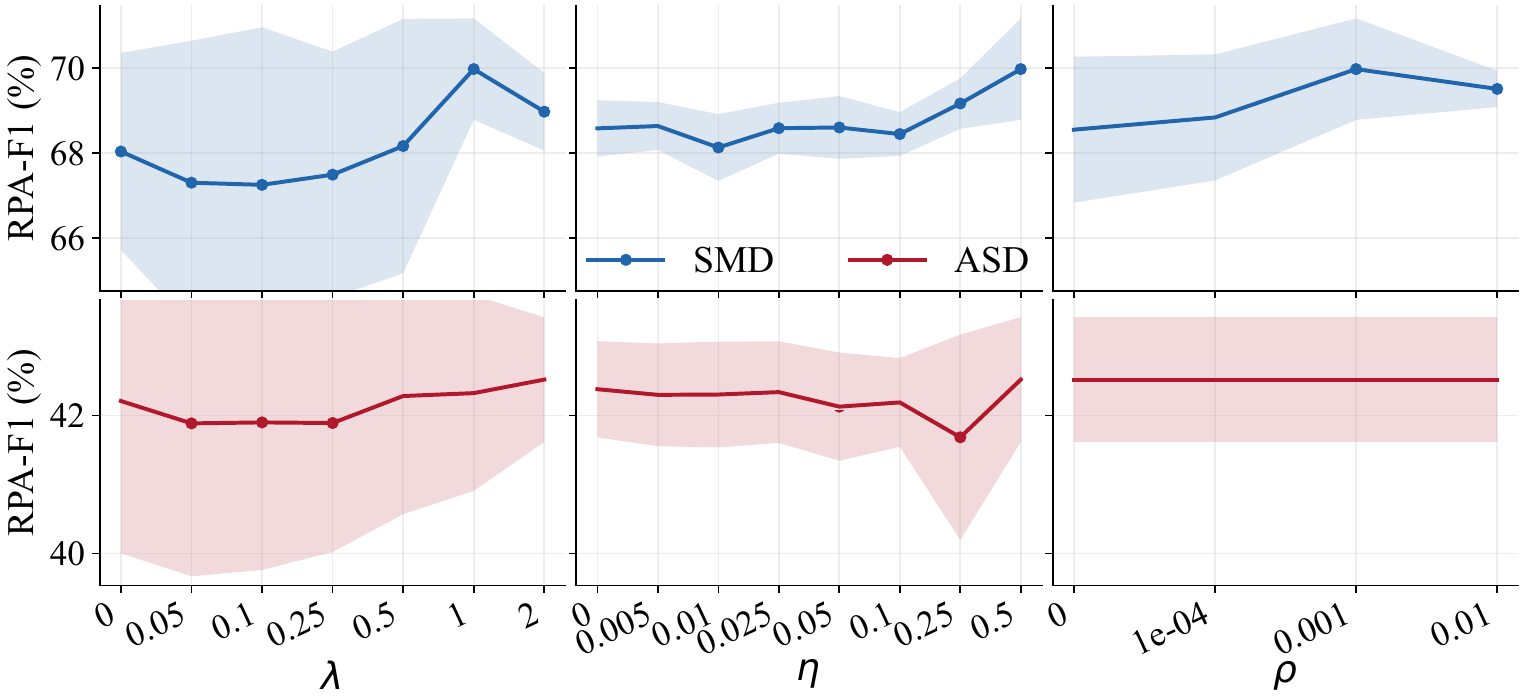}
    \caption{Sensitivity results of $\lambda$, $\eta$, and $\rho$ on SMD and ASD.}
    \label{fig:sense}
\end{figure}

\noindent\textbf{Training params.} Figure~\ref{fig:sense} studies three training params: contextual
reconstruction weight $\lambda$, probe supervision weight $\eta$, and the optional router-balance coefficient $\rho$. We report SMD and ASD as representative datasets with different degrees of temporal shift. On SMD, RPA-F1 is sensitive to all three coefficients: increasing $\lambda$ helps after a moderate value, consistent with contextual reconstruction being more valuable under pronounced temporal changes; $\eta$ also matters, supporting that heterogeneous probe supervision improves the router's assessment. SMD further benefits from a nonzero $\rho$, suggesting that a weak balance term can mitigate degenerate routing under strong non-stationarity. In contrast, ASD is much less sensitive, and the $\rho$ curve is nearly flat, implying that balance regularization is not a key gain source on relatively stable data. This contrast supports that the auxiliary terms mainly matter when temporal changes are prominent, while the model remains robust under milder shifts.

\subsection{4.5. Correction Behavior Analysis}
\begin{figure}[t]
    \centering
    \begin{minipage}[t]{0.48\linewidth}
        \centering
        \includegraphics[width=\linewidth]{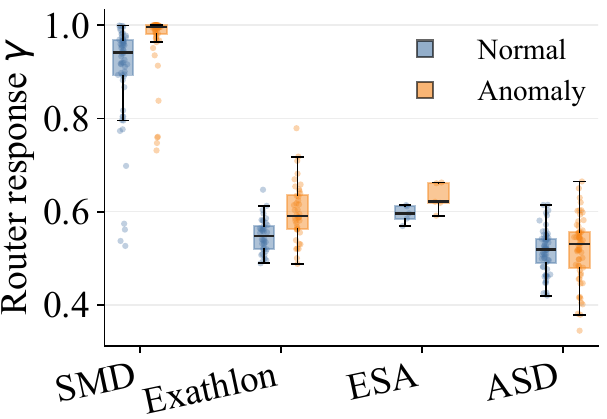}
        \caption*{(a) Router response.}
    \end{minipage}\hfill
    \begin{minipage}[t]{0.48\linewidth}
        \centering
        \includegraphics[width=\linewidth]{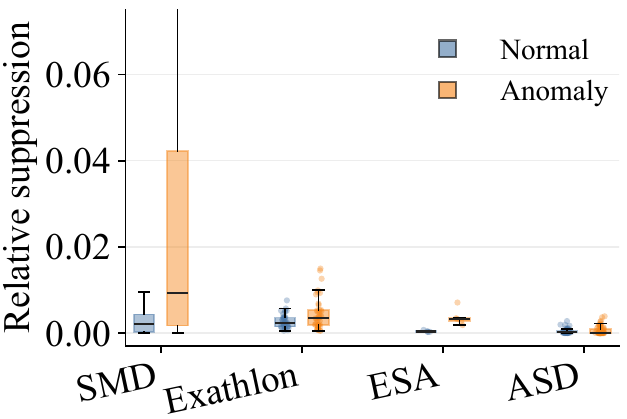}
        \caption*{(b) Score suppression.}
    \end{minipage}
    \caption{Correction on normal vs. anomalous windows. (a) Router response $\gamma_t$. (b) Relative score suppression from $\gamma_t$ and the positive reconstruction gap $\Delta_t$.}
\label{fig:router}
\end{figure}

\begin{figure}[ht]
    \centering
    \begin{minipage}[t]{0.48\linewidth}
        \centering
        \includegraphics[width=\linewidth]{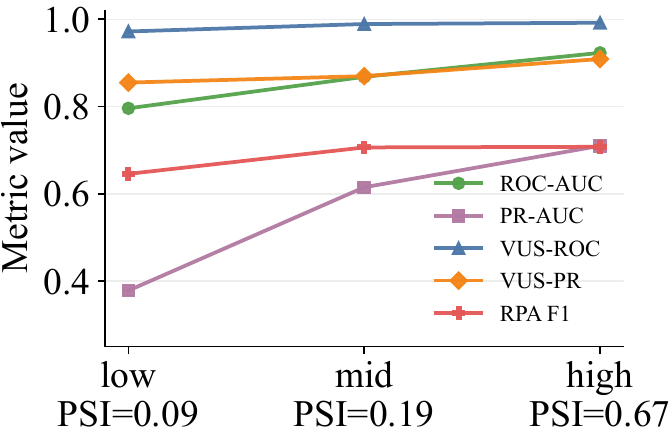}
        \caption*{(a) SMD.}
    \end{minipage}\hfill
    \begin{minipage}[t]{0.48\linewidth}
        \centering
        \includegraphics[width=\linewidth]{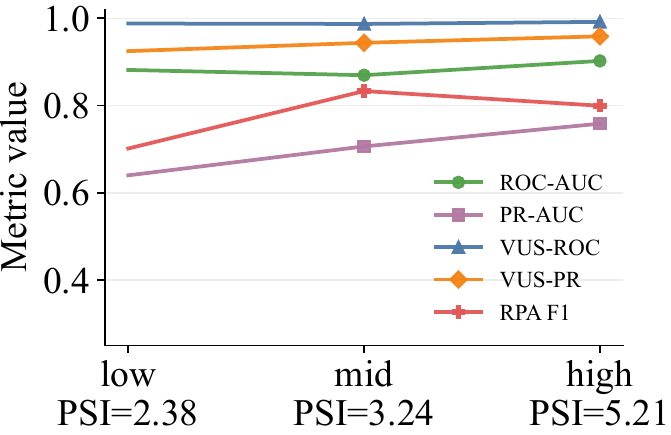}
        \caption*{(b) Exathlon.}
    \end{minipage}
    \caption{Detection performance of MORA across entity groups with different PSI levels.}
\label{fig:bucket}
\end{figure}

We inspect how MORA applies correction to normal and anomalous windows and how its performance varies with train–test shift intensity. 
Figure~\ref{fig:router} illustrates the two factors that govern score correction. The router response alone does not determine the attenuation strength. Instead, the realized correction is jointly determined by \(\gamma_t\) and the positive reconstruction gap \(\Delta_t\). For instance, although SMD yields high router responses, the resulting score suppression is modest and highly sample-dependent. Corrections are applied to both normal and anomalous windows because the router is not trained as a class predictor. Nevertheless, the one-sided rule constrains the corrected score to lie between the contextual and local reconstruction errors, preventing unconstrained score changes.
Figure~\ref{fig:bucket} presents a shift-stratified evaluation rather than a causal estimate of correction gains. On both SMD and Exathlon, detection performance remains stable or improves in higher-PSI groups, indicating that MORA does not suffer systematic degradation as the train--test distribution discrepancy increases.

\subsection{4.6. Visualization and Case Study }
\begin{figure}[h]
    \centering
    \includegraphics[width=0.85\columnwidth]{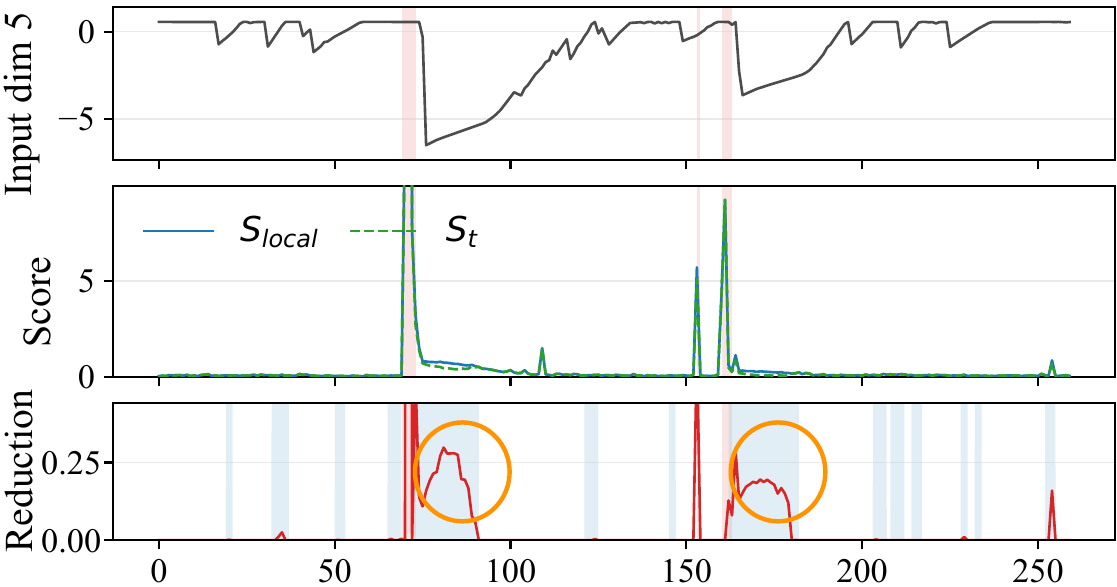}
    \caption{Visualization on SMD. From top to bottom: raw input, local and corrected scores, and score reduction $S_t^{loc}-S_t$. Red regions denote ground-truth anomalies, blue regions indicate non-anomalous intervals where correction is activated.}
    \label{fig:vis}
\end{figure}

Figure~\ref{fig:vis} provides a case study of MORA on SMD. Following the two labeled anomaly events near \(t\approx70\) and \(t\approx160\), the raw signal gradually recovers toward its preceding state. However, the local anomaly score remains elevated during parts of this recovery instead of decreasing promptly. MORA markedly attenuates these persistent post-event responses. Within the labeled intervals, the corrections remain small relative to the substantially larger local scores, particularly around \(t\approx160\), leaving the dominant anomaly peaks intact. This illustrates how broader temporal context reduces spurious responses without obscuring genuine anomaly evidence.

\section{Conclusion}
This work revisits drift-robust time-series anomaly detection as a problem of temporal change disambiguation: a local deviation should be judged not only against historical normality, but also in relation to its broader temporal evolution. MORA operationalizes this idea by reconstructing the same local target under paired short- and long-term views and translating the contextual reconstruction advantage into a one-sided correction of the primary local anomaly score. Heterogeneous cross-view cues further adapt the correction strength across samples, while the asymmetric scoring rule retains the local evidence whenever broader context provides no reconstruction improvement. Experiments on four TSAD benchmarks demonstrate strong and consistent performance, suggesting that temporal context is more effective when used as constrained corrective evidence than as an additional anomaly score to be fused. 
This perspective also motivates several extensions, including adaptive longer-range context selection, more ambiguity-aware correction routing, explicit modeling of shifts in cross-variable dependencies, and the integration of context-guided correction with broader TSAD backbones.

\bibliography{ref}


\end{document}